\documentclass[sigconf]{acmart}

\usepackage{hyperref}
\usepackage{multirow}
\usepackage[utf8]{inputenc}
\usepackage[ruled, lined, longend, linesnumbered]{algorithm2e}
\usepackage{amsmath,multicol}

\usepackage{amssymb,subcaption,textcomp,comment}

\usepackage{booktabs}
\usepackage{multirow}

\usepackage{array}
\newcolumntype{L}[1]{>{\raggedright\let\newline\\\arraybackslash\hspace{0pt}}m{#1}}
\newcolumntype{C}[1]{>{\centering\let\newline\\\arraybackslash\hspace{0pt}}m{#1}}
\newcolumntype{R}[1]{>{\raggedleft\let\newline\\\arraybackslash\hspace{0pt}}m{#1}}

\usepackage{titlesec}
\newcommand{\mwx}[1]{\textbf{\color{red}{#1}}}

\newcommand{\sys}{\texttt{MDLBench}\xspace}

\newcommand{\pytorchmobile}{\texttt{PyTorchMobile}\xspace}
\newcommand{\mnn}{\texttt{MNN}\xspace}
\newcommand{\ncnn}{\texttt{ncnn}\xspace}
\newcommand{\mace}{\texttt{Mace}\xspace}
\newcommand{\snpe}{\texttt{SNPE}\xspace}
\newcommand{\tflite}{\texttt{TFLite}\xspace}

\newcommand{\implication}[1]{$\dagger$\textbf{Implication} \textit{#1}}
\copyrightyear{2022}
\acmYear{2022}

\setcopyright{acmlicensed}\acmConference[WWW '22]{Proceedings of the ACM
Web Conference 2022}{April 25--29, 2022}{Virtual Event, Lyon, France.}
\acmBooktitle{Proceedings of the ACM Web Conference 2022 (WWW '22), April 25--29, 2022, Virtual Event, Lyon, France}
\acmPrice{15.00}
\acmISBN{978-1-4503-9096-5/22/04}
\acmDOI{10.1145/3485447.3512148}
\begin{document}

	\title{A Comprehensive Benchmark of Deep Learning Libraries on Mobile Devices}

	\author{Qiyang Zhang$^1$, Xiang Li$^2$, Xiangying Che$^1$, Xiao Ma$^1$, Ao Zhou$^1$, Mengwei Xu$^1$}
	\authornote{Corresponding author: Mengwei Xu (mwx@bupt.edu.cn).}
	\author{Shangguang Wang$^1$,Yun Ma$^3$, Xuanzhe Liu$^3$}
	
	\affiliation{%
	\institution{Beijing University of Posts and Telecommunications$^1$,China}
		\country{}
	}
	\affiliation{%
		\institution{China University of Petroleum$^2$,China}
		\country{}
	}
	\affiliation{%
		\institution{Peking University$^3$,China}
		\country{}
	}

	\renewcommand{\shortauthors}{Qiyang Zhang et al.}

	\begin{abstract}
		Deploying deep learning (DL) on mobile devices has been a notable trend in recent years.
		To support fast inference of on-device DL, DL libraries play a critical role as algorithms and hardware do.
		Unfortunately, no prior work ever dives deep into the ecosystem of modern DL libs and provides quantitative results on their performance. In this paper, we first build a comprehensive benchmark that includes 6 representative DL libs and 15 diversified DL models.
		We then perform extensive experiments on 10 mobile devices, which help reveal a complete landscape of the current mobile DL libs ecosystem.
		For example, we find that the best-performing DL lib is severely fragmented across different models and hardware, and the gap between those DL libs can be rather huge.
		In fact, the impacts of DL libs can overwhelm the optimizations from algorithms or hardware, e.g., model quantization and GPU/DSP-based heterogeneous computing.
		Finally, atop the observations, we summarize practical implications to different roles in the DL lib ecosystem.
		
	\end{abstract}



	\maketitle

	\section{Introduction}\label{sec:intro}

	\begin{table}[t]
		\footnotesize
		\begin{tabular}{|l|L{1cm}|L{2.4cm}|L{1.2cm}|}
			\hline
			\textbf{Benchmark} & \textbf{Scenario} & \textbf{Benchmark objective} & \textbf{Supported DL libs} \\ \hline
			MLPerf~\cite{mattson2019mlperf} & T/I@S/E & Hardware   & /  \\ \hline
			DeepBench~\cite{DeepBench} & T/I@S/E & Hardware  & / \\ \hline
			DAWNBench~\cite{DAWNBench} & T/I@S & Hardware, algorithm, and DL libs & / \\ \hline
			AI Matrix~\cite{AI-Matrix} & I@S & Hardware and DL libs & 4 \\ \hline
			AI-Benchmark~\cite{ignatov2019ai} & I@E & Hardware & 1 \\ \hline
			Fathom~\cite{adolf2016fathom} & T/I@S & Algorithm & 1 \\ \hline
			gaugeNN~\cite{almeida2021smart} & I@E & DL apps and models & 1 \\ \hline
			AIIA~\cite{AIIA} & I@E & Hardware & 3 \\ \hline
			This work & I@E & DL libs & 6 \\ \hline 
		\end{tabular}  
	\caption{Comparison of this work and existing AI benchmarks. ``T/I'': training/inference stages; ``S/E'': server/edge platform. ``/'' means the benchmark ask the third-party users to submit their own software.}
	\label{tab:DL_libs}
	\vspace{-15pt}
	\end{table} 
Deep learning has become a key enabler towards ubiquitous and intelligent web applications like Web AR~\cite{yang2020subsequent,deeptype,ren2021fine,qiao2018new,liu2014imashup}.
A noteworthy trend is that more and more Deep learning inference tasks are now shifting from cloud datacenters to smartphones, making a case for low user-perceived delay and data privacy preservation with the support of on-device DL~\cite{qiao2018new}.
For example, it is reported that the DL-embedded apps on Google Play market have increased by 60\% from Feb. 2020 to Apr. 2021, and those apps contribute to billions of downloads and user reviews~\cite{almeida2021smart, xu2019first}.

Running inference (or prediction) task in a fast way is the intuitively basic demand to on-device DL, as many of them are deployed for continuous user interactions.
It is also fundamentally challenging because DL models are known to be very complex and cumbersome~\cite{almeida2019embench,huang2017densely,pratap2020scaling}.
Consequently, optimizing the inference performance has been the theme of both academia~\cite{deeptype,xu2018deepcache,almeida2021smart} and industry~\cite{market,funf,AImarket} in recent years.

The inference performance of on-device DL is affected by many factors.
Existing literature that aims to quantitatively understand the performance mostly focuses on \underline{hardware} and \underline{models}, leaving the \underline{software} (DL execution engines or \emph{DL libs}) underexplored. Yet, software also plays a critical role in speeding up the on-device DL inference, e.g., up to 62,806$\times$ gap between vanilla and fine-tuned implementation~\cite{leiserson2020there}.

Furthermore, due to the severely fragmented ecosystem of smartphones~\cite{DBLP:journals/tse/WeiLCHLL20}, there exists a mass of heterogeneous DL lib alternatives for app developers~\cite{almeida2021smart,xu2019first}, making it difficult and labor-intensive to compare their suitability into specific models.

To gain in-depth understandings of the performance of modern DL libs, we first build a comprehensive benchmark for on-device DL inference, namely \sys.
The benchmark includes 6 popular, representative DL libs on mobile devices, i.e., \tflite, \pytorchmobile, \ncnn, \mnn, \mace, and \snpe~\cite{tensorflow-performance,pytorchm,ncnn,mnn,Mace,SNPE}.
It contains 6 DL models compatible with all above DL libs and 8 models compatible with at least 3 above DL libs, spanning from image classification, object detection, to NLP.
Compared to existing AI benchmarks (Table~\ref{tab:DL_libs}), our benchmark triumphs at the aspect of rich support for various DL libs and models.
In addition to the completeness, we also instrument the DL libs to obtain underlying performance details such as per-operator latency, CPU usage, etc.
Those details allow us to peek into the intrinsic features of those DL libs and therefore provide more insightful implications to developers and contribute to a more useful, and less fragmented Web of Things.

Based on our benchmark, we perform extensive experiments to demystify the performance of DL libs on various models (15 in total) and hardware (10 smartphones that are equipped with CPU/GPU/DSP).
Through the experiments, we make the following interesting and useful observations.

\textbf{(1) The performance of the 6 DL libs benchmarked is severely fragmented across different models and hardware ($\S$\ref{sec:fragmentation}).}
There is no \textbf{One-Size-Fits-All} DL lib that performs best on all scenarios (model$\otimes$device), yet each DL lib has at least one best-performing scenario.
Even for the same model, there are different DL libs that perform the best on different devices.
The performance gap between those DL libs is huge, i.e., about 7.4$\times$/1.9$\times$ among the best and the worst/2nd-best ones on average.
On mobile GPU, such fragmentation is further exaggerated by the multiple choices of software driver backends (Vulkan/OpenGL/OpenCL).

\textbf{(2) The impacts of DL software may overwhelm the algorithm designs and hardware capacity ($\S$\ref{sec:quantization}, $\S$\ref{sec:accelerator}).}
Designing a more lightweight model structure, model quantization (FP32 to INT8), and using mobile GPUs/DSPs with high parallelism are common ways to speed up on-device inference.
However, due to the defects from DL libs, those methods cannot always bring expected benefits or even slow down the inference.
For instance, INT8-based quantization only brings 0.8$\times$--3.0$\times$ inference speedup ($\le$1 means slowdown), which is much less than the theoretical expectation, i.e., 4$\times$ due to the NEON support in Android~\cite{jo2014opencl}.

\textbf{(3) There is a noteworthy potential to further enhance the DL lib performance by integrating the optimal implementation of different DL libs at operator level ($\S$\ref{sec:optimal}).}
Motivated by the severe fragmentation of DL libs, we perform an emulation test assuming that we can combine the best of each DL libs at operator level.
Through such combinations, we can potentially reduce the inference time by up to 29.9\% compared to the best-performing lib.

\textbf{(4) Cold-start inference of DL libs is significantly slower than warm inference ($\S$\ref{sec_design_coldstart})}.
On average, the first inference for each session is about 10.8$\times$ and 25.7$\times$ slower than the following ones on CPU and GPU, respectively.
Diving deeper, we find that the memory preparation stage contributes to the most of the overhead, which includes expanding the loaded weights to proper memory locations and reserving memory for intermediate feature maps.

\textbf{(5) During the evolution of DL libs, performance bugs are introduced for many times ($\S$\ref{sec:long}).}
By benchmarking the weekly version of \tflite and \ncnn since early 2018, we find that the overall performance of DL libs is improving yet becomes relatively stable since 2020.
Surprisingly, we observe that some commits incur significant performance degradation on certain scenarios, which can take 1--16 weeks to be fixed.
For example, some commits provide new forms of operator implementation, aimed to improve the inference performance.
Those changes, however, result in performance degradation in certain models or devices.

Atop the above observations, we summarize the key implications to different roles in the mobile DL ecosystem.

\textbf{\textit{For DL app/model developers}}: (i) It is extremely challenging in selecting a proper DL lib due to the severe fragmentation.
To pursue the optimal performance under each scenario, they have to embed different DL libs into the apps and load one dynamically based on the model and hardware settings.
(ii) A more lightweight model (fewer FLOPs) does not always run fast.
The impact from software at deployment needs to be considered during model design.

\textbf{\textit{For DL lib engineers and researchers}}: (i) It is time to review the pros and cons of different DL libs and work out a solution that can integrate their wisdom in a unified manner.
Otherwise, the fragmentation may continuously exist for a long term as fixing it can take huge amount of engineering efforts.
(ii) The cold-start inference time is a rarely touched topic, but can be important in apps that only need to execute models for one time per session.
Potential optimizations include using multi-thread to speed up memory preparation and operator-level pipeline of different initialization stages.
(iii) Performance bugs bring negative impacts to the open-source ecosystem of DL libs, but are difficult to be fully eliminated due to the aforementioned fragmentation.
Tools that can automatically identify such bugs timely, either through dynamic or static analysis, are urgently needed.

Our main contributions are as follows.
\begin{itemize}
\item We design and implement \sys, a fully automatic, comprehensive benchmark for DL libs. 
The full benchmark suite is open-sourced\footnote{https://github.com/UbiquitousLearning/MobileDLFrameworksBenchmark}.

\item We perform extensive experiments with \sys on diverse hardware devices and models, and the results reveal a complete landscape of the current DL lib ecosystem.

\item We summarize the insightful observations and practical implications based on the experiments that can benefit different roles in the DL lib ecosystem.

\end{itemize}
\section{Benchmark and Methodology}\label{sec:bkgnd}

\begin{table*}[htpb]
	\small
	\begin{tabular}{|l|l|l|l|l|l|l|l|}
	\hline
	 \textbf{Models } & \textbf{Tasks} & \textbf{TFLite } & \textbf{ncnn} &\textbf{mnn}  & \textbf{MACE} & \textbf{PyTorchMobile} & \textbf{SNPE}    \\ \hline
	mobilenetV1\cite{mobilenet} & image classification     & $C_{32,8}$-$G_{32,8}$-$D_{8}$ & $C_{32,8}$-$G_{32,8}$ & $C_{32,8}$-$G_{32,8}$ & $C_{32,8}$-$G_{32}$ & $C_{32,8}$    & $C_{32,8}$-$G_{32,8}$-$D_{8}$    \\ \hline
	mobilenetV2\cite{sandler2018mobilenetv2} &  image classification    & $C_{32,8}$-$G_{32,8}$-$D_{8}$ & $C_{32,8}$-$G_{32,8}$ & $C_{32,8}$-$G_{32,8}$ & $C_{32,8}$-$G_{32}$ & $C_{32,8}$    & $C_{32,8}$-$G_{32,8}$-$D_{8}$   \\ \hline
	inceptionV3  \cite{inceptionv3} & image classification     & $C_{32,8}$-$G_{32,8}$-$D_{8}$ & $C_{32,8}$-$G_{32,8}$ & $C_{32,8}$-$G_{32,8}$ & $C_{32}$-$G_{32}$ & $C_{32,8}$    & $C_{32,8}$-$G_{32,8}$-$D_{8}$ \\ \hline
	inceptionV4  \cite{inceptionv4} & image classification     & $C_{32,8}$-$G_{32,8}$-$D_{8}$ & $C_{32,8}$-$G_{32,8}$ & $C_{32}$-$G_{32}$     & $C_{32}$-$G_{32}$ & $C_{32,8}$    & $C_{32,8}$-$G_{32,8}$-$D_{8}$  \\ \hline
	vgg16  \cite{vgg16}  & image classification         & $C_{32,8}$-$G_{32,8}$-$D_{8}$ & $C_{32,8}$-$G_{32,8}$ & $C_{32,8}$-$G_{32,8}$ & $C_{32}$-$G_{32}$ & $C_{32,8}$    & $C_{32,8}$-$G_{32,8}$-$D_{8}$  \\ \hline
	squeezenet  \cite{squeezenet}& image classification      & $C_{32,8}$-$G_{32,8}$-$D_{8}$ & $C_{32,8}$-$G_{32,8}$ & $C_{32}$-$G_{32}$     & $C_{32}$-$G_{32}$ & $C_{32,8}$    & $C_{32,8}$-$G_{32,8}$-$D_{8}$ \\ \hline
	nasnet\_mobile \cite{nasnet} & image classification  & $C_{32}$-$G_{32}$     & -                     & $C_{32}$-$G_{32}$     & $C_{32}$-$G_{32}$ & $C_{32}$      & -                    \\ \hline
	densenet   \cite{densenet} &  image classification       & $C_{32}$-$G_{32}$     & -                     & $C_{32}$-$G_{32}$     & -                 & $C_{32}$      & $C_{32}$-$G_{32}$  \\ \hline
	mnasnet   \cite{mnasnet}   &  image classification      & $C_{32}$-$G_{32}$     & $C_{32}$-$G_{32}$     & $C_{32}$-$G_{32}$     & $C_{32}$-$G_{32}$ & $C_{32}$      & $C_{32}$-$G_{32}$   \\ \hline
	resnetv2\_50  \cite{resnetv2}& image classification    & $C_{32}$-$G_{32}$     & $C_{32}$-$G_{32}$     & $C_{32}$-$G_{32}$     & $C_{32}$-$G_{32}$ & $C_{32}$      & $C_{32}$-$G_{32}$    \\ \hline
	deeplabv3    \cite{deeplabv3} &  semantic segmentation    & $C_{32}$-$G_{32}$     & -                     & $C_{32}$-$G_{32}$     & $C_{32}$-$G_{32}$ & -             & -        \\ \hline
	ssd\_mobilenetV1 \cite{ssd} &  object detection & $C_{32}$-$G_{32}$    & $C_{32}$-$G_{32}$     & $C_{32}$-$G_{32}$     & $C_{32}$-$G_{32}$ & $C_{32}$      & -                 \\ \hline
	yolo-fastest  \cite{yolofaster}   &  object detection  & $C_{32}$-$G_{32}$     & $C_{32}$-$G_{32}$     & $C_{32}$-$G_{32}$     & -                 & -             & -      \\ \hline
	yolo3    \cite{yolov3}     &  object detection     & $C_{32}$-$G_{32}$     & $C_{32}$-$G_{32}$     & $C_{32}$-$G_{32}$     & -                 & -             & -    \\ \hline
	albert\_tiny \cite{albert}   &    text classification   & $C_{32}$-$G_{32}$     & -                     & $C_{32}$-$G_{32}$     & -                 & -             & -     \\ \hline
	\end{tabular}
	\caption{The supported DL libraries and models of \sys. ``C/G/D'': mobile CPU/GPU/DSP. The subscripted 32 and 8 represent different model precision, i.e., float32 and int8, respectively.}
	\label{tab:libs}
	\vspace{-15pt}
	\end{table*}

\sys is a benchmark aimed to understand the impacts of DL libs on the on-device DL performance.
It has the following advantages over existing AI benchmarks.

\noindent $\bullet$ \textbf{Rich support}
Table~\ref{tab:libs} summarizes the DL libs (6 in total), models (15 in total), and hardware processor (CPU/GPU/DSP) \sys currently supports.
Being able to test many DL libs under various contexts is critical to obtain a complete landscape of the DL lib ecosystem, because the performance optimization is quite ad-hoc to models and hardware.
Among the large amount of DL libs available for developers, we select 6 most representative ones from a ``market'' perspective.
We follow the prior works~\cite{xu2019first} to detect the DL libs used in 16,000 Android apps we crawled in Mar. 2021 from Google Play.
Among the 676 DL apps identified, we find the most popular DL libs are TFLite (70.5\%), TensorFlow (7.8\%), ncnn (7.2\%), caffe (4.4\%), MNN (4.1\%), PyTorchMobile (3.8\%), Mace (1.2\%).
We filter TensorFlow and caffe, as their support for smartphones are deprecated a few years ago and has been merged into the corresponding lightweight implementation, i.e., TFLite and PyTorchMobile.
We further include SNPE into \sys, as it's a vendor-specific (Qualcomm) DL lib while all above are not.
The models we use come from two sources.
One is the model zoo of TensorFlow and PyTorch~\cite{tfmo,pymo}.
The other is by using the built-in converters of each DL lib to convert models to different formats~\cite{ncnn,Mace,mnn,SNPE}.
\sys also incorporates a module to automatically check the equivalence of the same model generated for different DL libs.

\noindent $\bullet$ \textbf{Detailed metrics}
\sys profiles the inference time and operator-level information, e.g., per-operator latency, duration, input/output dimension, etc.
Such functionality is originally supported for some DL libs (TFLite, SNPE) yet for others (MNN) we need instrumenting the source code.
\sys then adds another layer of traces processing to unify and visualize the output from different DL libs.


\begin{figure}[t]
	\centering
	\begin{minipage}[b]{0.46\textwidth}
		\includegraphics[width=1\textwidth]{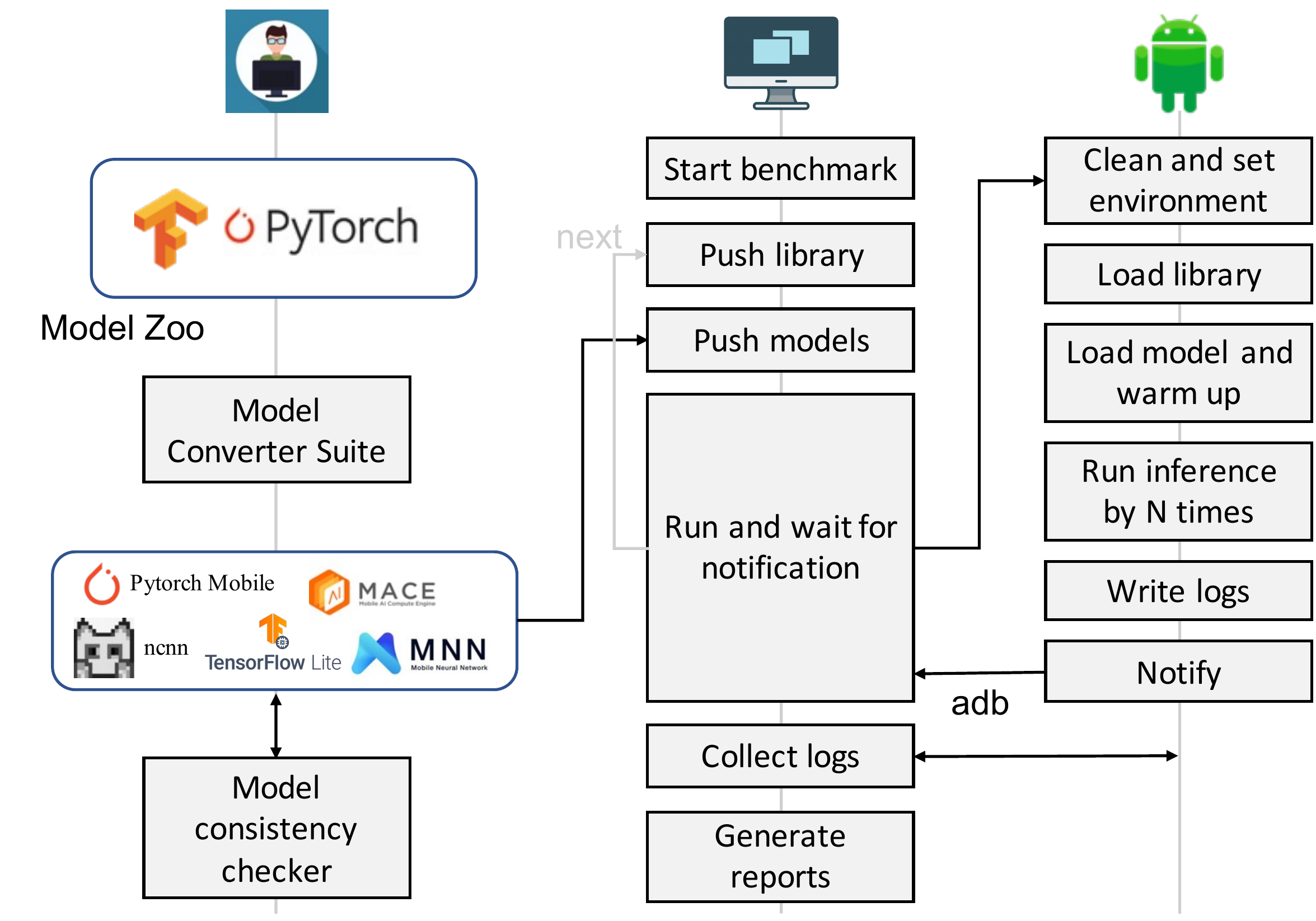}
	\end{minipage}
	\caption{Workflow of \sys.}
	\label{fig:workflow}
	\vspace{-15pt}
\end{figure}
\textbf{Workflow} Figure~\ref{fig:workflow} shows the overall workflow of \sys.
For each testing, the desktop-side benchmark iterates over each DL lib.
It first pushes the lib and corresponding models generated as aforementioned to the devices through adb~\cite{kim2013study}.
Next, the device cleans the system environment by killing other apps in background and sets the system configurations (CPU frequency, thread number, etc).
Following prior work~\cite{wang2021asymo,cai2021towards}, we always use 4 big cores to run the DL libs as it's often the best-performing setting.
The device then loads the library and model into memory to warm up, and executes the inference by N times (50 by default).
The testing results will be written to device storage and retrieved to desktop.

\textbf{Devices}
Table~\ref{tab:devices} shows the devices used in our measurement.
It includes 10 different device models with various SoCs (Snapdragon series, Kirin, Helio) and GPUs (Adreno series and Mali series),
where the currently selected smartphones are still representative to reflect the hardware heterogeneity.
\begin{table}[htpb]
	\small
	\begin{tabular}{|l|l|l|l|l|l|}
	\hline
	\textbf{Devices}         & \textbf{abbr.} & \textbf{SoC} & \textbf{GPU}   & \textbf{RAM} \\ \hline 
	Google Pixel5   & GP5   & Snapdragon 765G & Adreno 620    & 8GB \\ \hline 
	Huawei Enjoy 8  & HE8   & Snapdragon 430  & Adreno 505    & 4GB \\ \hline
	MeiZu 16T       & MZ16  & Snapdragon 855  & Adreno 640    & 6GB \\ \hline
	HuaWei Mate30   & HM    & Kirin 990       & Mali-G76 MP16 & 8GB \\ \hline 
	XiaoMi11 Pro    & MI11  & Snapdragon 888  & Adreno 660    & 8GB \\ \hline 
	XiaoMi9         & MI9   & Snapdragon 855  & Adreno 640    & 6GB \\ \hline 
	MeiZu 16T       & MZ16  & Snapdragon 855  & Adreno 640    & 6GB \\ \hline
	OnePlus 9R      & OP9   & Snapdragon 870  & Adreno 650    & 8GB \\ \hline 
	RedMi9          & R9    & Helio G80       & Mali-G52 MC2  & 4GB \\ \hline 
	Redmi Note9 Pro & RN9   & Snapdragon 720G & Adreno 618    & 6GB \\ \hline 
	Samsung S21     & S21   & Snapdragon 888  & Adreno 660    & 8GB \\ \hline                          
	\end{tabular}
	\caption{The tested devices and their specifications.}
	
	\label{tab:devices}
	\vspace{-15pt}
	\end{table}
	\section{Performance Analysis}\label{sec:design}

	Based on \sys and the diverse mobile devices, we perform extensive experiments and analyze the results.
	The theme of this measurement is to quantitatively understand the performance discrepancy of different DL libs, and how the inter-play with the impacts from algorithm and hardware.
	Besides, we also explore two rarely touched topics in mobile community: what is the performance of the first inference (cold start) of different DL models, and how does the performance of DL libs evolve across time.

\subsection{Performance Fragmentation}\label{sec:fragmentation}

\begin{figure}[t]
	\centering
	\begin{minipage}[b]{0.45\textwidth}
		\includegraphics[width=1\textwidth]{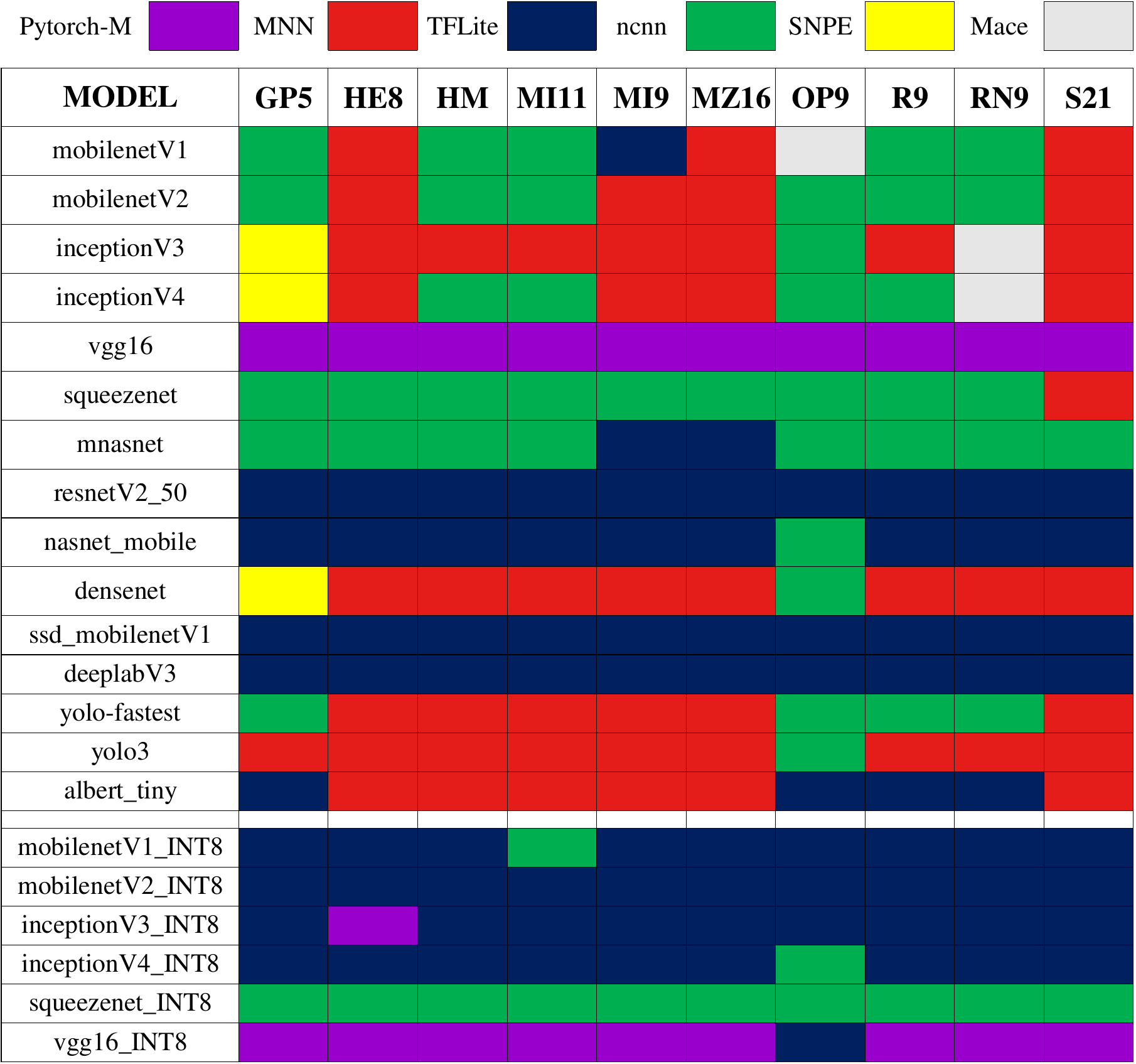}
		\subcaption{CPU}
	\end{minipage}

	\begin{minipage}[b]{0.45\textwidth}
		\includegraphics[width=1\textwidth]{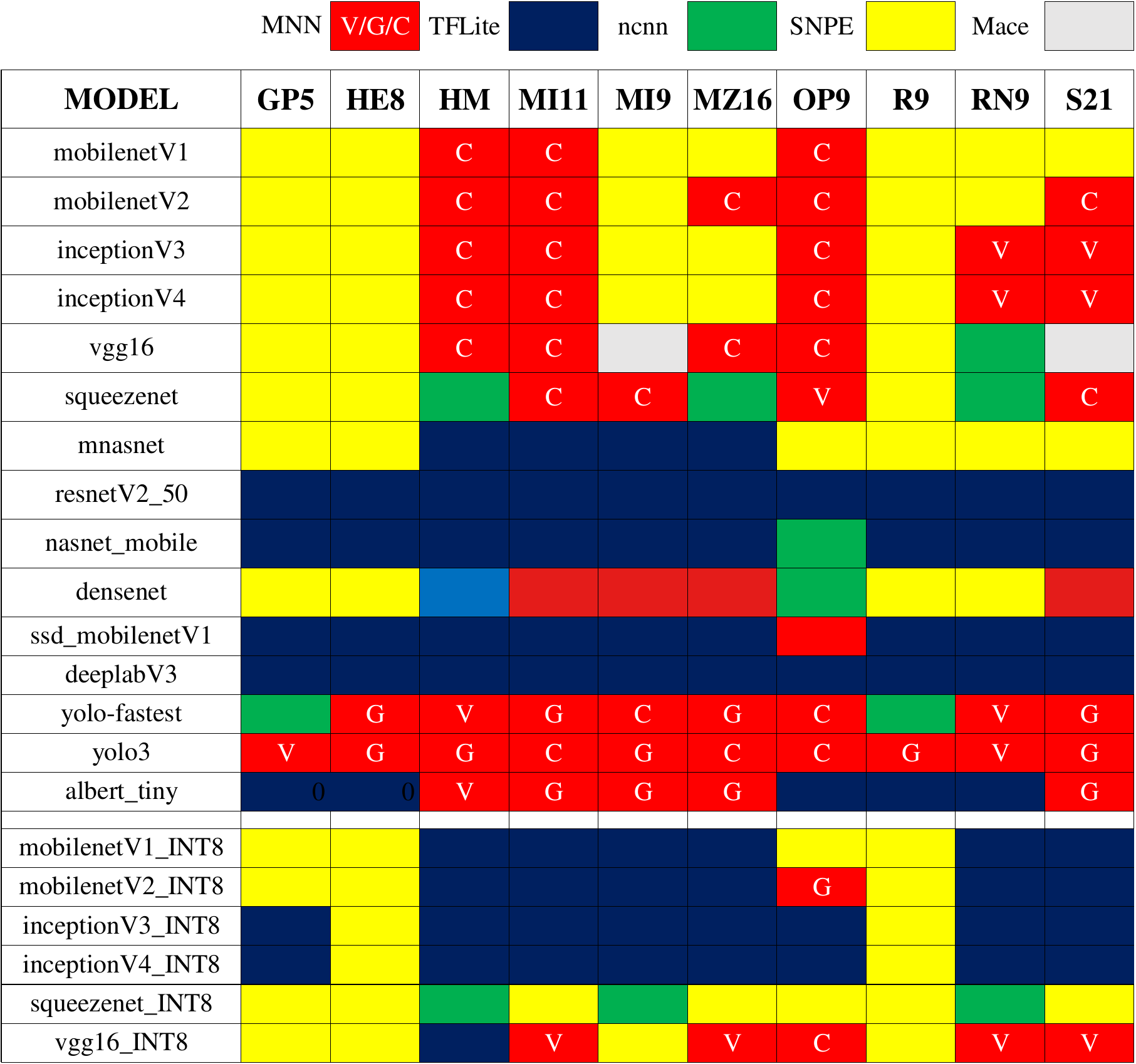}
		\subcaption{GPU. The characters in \mnn indicate different GPU backends: V/G/C indicate Vulkan/OpenGL/OpenCL.}
	\end{minipage}
	
	\caption{The best-performing DL lib (smallest inference time) with different models and devices.}
	
	\label{fig:heatmap}
	\vspace{-10pt}
\end{figure}

\begin{figure}[t]
	\centering
	\begin{minipage}[b]{0.4\textwidth}
		\includegraphics[width=1\textwidth]{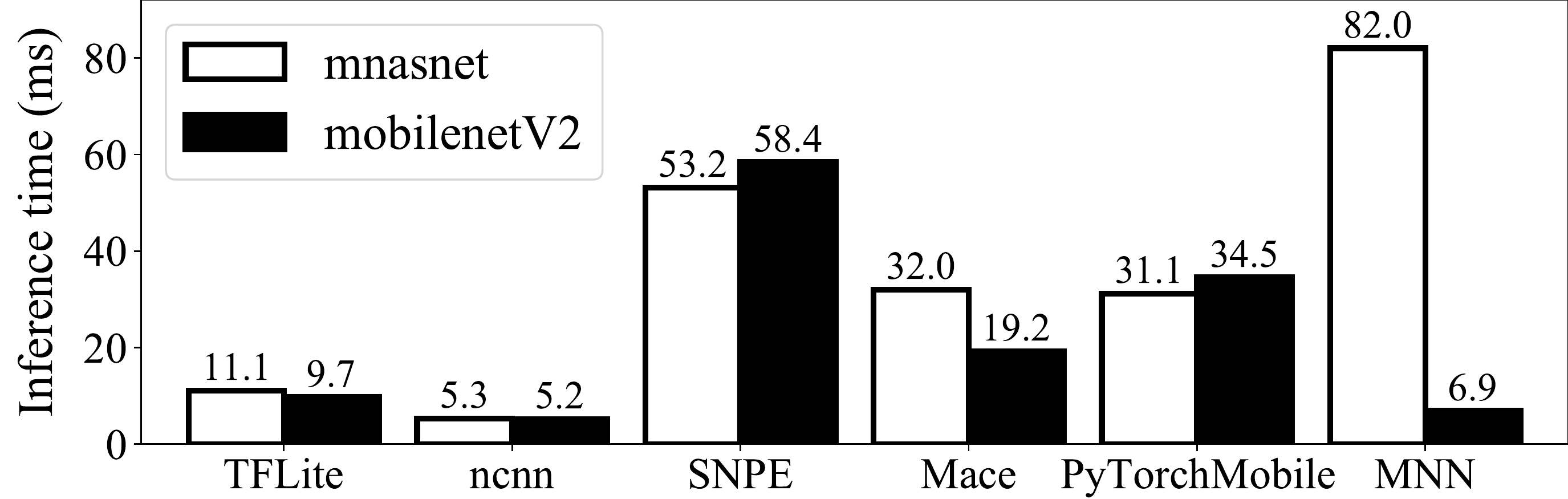}
	\end{minipage}
	\caption{The average inference time of squeezenet and mnasnet with different libs on MI11.}
	\label{fig:model}
	\vspace{-15pt}
\end{figure}
Figure~\ref{fig:heatmap} summarizes the best-performing DL lib (by color), i.e., the DL lib with the smallest inference time when running different models on heterogeneous devices.
We observe that the performance of DL libs across models and hardware devices is severely fragmented.

(1) \textbf{There is no one-size-fit-all DL lib for optimal performance across models and hardware.}
On either CPU and GPU, a silver-bullet best-performing DL lib does not exist.
Different DL libs have different adept scenarios (model $\times$ device), Table~\ref{tab:best-perf} showcases the number of best performing scenarios
of different DL libs, as summarized from Figure~\ref{fig:heatmap}.
Each DL lib has at least one best-performing scenario, except that \pytorchmobile does not support GPU acceleration.
Even for the same model, its corresponding best-performing DL lib may change across different hardware.
For instance, the best-performing DL libs of inceptionV3 are \snpe, \ncnn, and \mace on GP5, OP9, and RN9, respectively.

Such high performance fragmentation mainly attributes to two facts. First, mobile hardware ecosystem is highly fragmented in consideration of their Big.Little Core architecture, cache size, GPU capacity, etc.
Second, the model structure is also heterogeneous. Implementing depth-wise convolution operator~\cite{guo2019depthwise} is totally different from traditional convolution operators as they have different cache access patterns.
Overall, the fragmentation of models and hardware forces the software-level DL inference optimization especially model- and hardware-specific.
To obtain the optimal performance, DL lib developers need to hand-craft each operator at very low-level programming interface, heavily relying on assembly language and NEON instructions.
While being able to fully exploit the capacity of specific hardware, such implementation cannot be generalized well to different hardware platforms.
For example, \ncnn has 44 different types of implementation for convolution operation, each fitting to different execution contexts like kernel size, hardware architecture, etc.
Due to the high manual programming efforts, there is no oracle DL lib optimized for each scenario.
\begin{table}[t]
	\small
	\begin{tabular}{|l|r|r|r|r|r|r|}
		\hline
		& \textbf{\tflite} & \textbf{\pytorchmobile} & \textbf{\ncnn} & \textbf{\mnn} & \textbf{\mace} & \textbf{\snpe} \\ \hline
		\textbf{CPU} & 84 & 20 & 52 & 48 & 3 & 3 \\ \hline
		\textbf{GPU} &80 & / & 12 & 49 & 5 & 64 \\ \hline
	\end{tabular}
\caption{The number of best-performing scenario of each DL lib, summarized from Figure~\ref{fig:heatmap}. A scenario is defined by a pair of DL model and a tested device.}
\label{tab:best_perf}
\vspace{-15pt}
\end{table}

(2) \textbf{The performance gap of DL libs can be large.}
To show the absolute numerical gap between DL libs performance,
Table~\ref{tab:subs} summarizes the performance gap of 6 models between the best-performing DL libs and the worst/2nd-best.
The "gap" is defined as the ratio of inference time of two DL libs (the longer one divided by the shorter one).
The numbers in parentheses are average values.
Surprisingly, though those DL libs are all specifically designed and optimized for mobile devices, the performance gap can be quite severe.
For instance, for the same model vgg16, the gap between different libs and smartphones is as high as 54.3$\times$, and even the smallest gap between the best and the second best is 1.5$\times$.
On average, the gap between the best-performing to the worst one is 7.4$\times$, and to the 2nd-best one is 1.9$\times$.

\begin{table}[t]
    \small
    \begin{tabular}{|c|c|c|c|c|c|}
        \hline
        \multicolumn{1}{|c|}{\multirow{2}{*}{\textbf{Models}}} & \multicolumn{2}{c|}{\textbf{Best vs. Worst}}               & \multicolumn{2}{c|}{\textbf{Best vs. 2nd Best}}   \\ 
                                                     & \textbf{CPU ($\times$) }                  & \textbf{GPU ($\times$) }                 & \textbf{CPU ($\times$) }                & \textbf{GPU ($\times$)  }               \\ \hline
        mobilenetV1                                  & 4.0\textasciitilde15.4 (8.7)  & 1.7\textasciitilde14.1 (5.6) & 1.1\textasciitilde1.9 (1.5) & 1.0\textasciitilde4.0 (1.9) \\ \hline
        mobilenetV2                                  & 5.6\textasciitilde18.8 (11.2) & 2.9\textasciitilde15.9 (6.2) & 1.1\textasciitilde2.0 (1.5) & 1.0\textasciitilde2.9 (1.6) \\ \hline
        inceptionV3                                  & 2.6\textasciitilde5.6 (3.8)   & 3.0\textasciitilde13.4 (7.1) & 1.1\textasciitilde2.4 (1.7) & 1.0\textasciitilde4.0 (2.1) \\ \hline
        inceptionV4                                  & 2.0\textasciitilde5.4 (3.2)   & 2.4\textasciitilde11.0 (5.8) & 1.1\textasciitilde2.0 (1.5) & 1.0\textasciitilde3.6 (2.0) \\ \hline
        \multicolumn{1}{|c|}{vgg16}                  & 7.1\textasciitilde54.3 (16.2) & 4.4\textasciitilde7.0 (5.5)  & 1.3\textasciitilde4.2 (2.4) & 1.1\textasciitilde2.2 (1.5) \\ \hline
        squeezenet                                   & 4.6\textasciitilde19.9 (9.1)  & 1.9\textasciitilde12.6 (5.9) & 1.0\textasciitilde5.9 (2.5) & 1.1\textasciitilde2.5 (1.6) \\ \hline
        \underline{average}                                      &              \underline{8.7}                &            \underline{6.0}                   &              \underline{1.9}               &          \underline{1.8}     \\ \hline
    \end{tabular}
    \caption{The performance gaps of different DL libs.}
    \label{tab:subs}
    \vspace{-15pt}
\end{table}
\begin{figure*}[t]
	\centering
	\begin{minipage}[b]{0.8\textwidth}
		\includegraphics[width=1\textwidth]{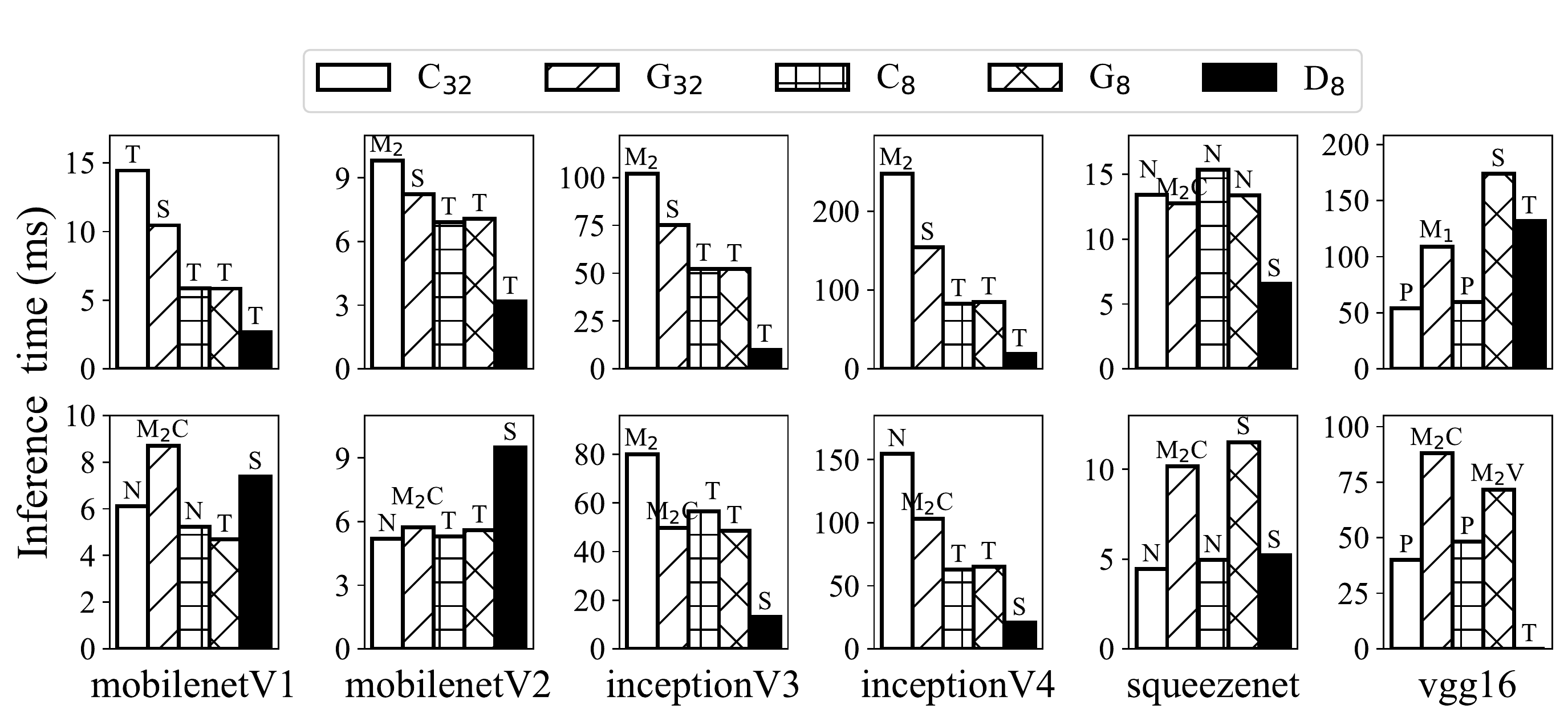}
	\end{minipage}
	\caption{The best inference speed across all DL libs. "T", "N", "S", "P", "M$_{1}$", and "M$_{2}$" are short for
	\tflite, \ncnn, \snpe, \pytorchmobile, \mace, and \mnn respectively as the best-performing DL libs. "V", "C", "G" indicate different GPU backends. We leave out vgg16 with SNPE since the model does not work on MI11.}
	\label{fig:processor_compare}
	\vspace{-10pt}
\end{figure*}

(3) \textbf{GPU backend choices further exaggerate the fragmentation.}
Even on the same GPU, there are different backend choices implemented by DL libs.
For example, \mnn implements three backends: Vulkan, OpenGL and OpenCL~\cite{sellers2016vulkan,mues2020optimization,jo2014opencl}.
Interestingly, as shown in Figure~\ref{fig:heatmap}(b), different GPU backend choices also fit different models and devices.
This is somehow surprising because Vulkan in \mnn is mainly used for cross-platform compatibility (e.g., desktop), while OpenGL/OpenCL are mobile-specific programming interfaces highly optimized for mobile devices~\cite{mues2020optimization}.
Such phenomenon attributes to both the underlying design of backends and how DL developers implement the DL operators atop the backends.

(4) \textbf{With software heterogeneity, the model structure is not the sole factor that determines their relative performance.}
We deem that model complexity does affect the inference time, e.g., the computation complexity represented by floating-point operations (FLOPs) and the number of models parameters.
In fact, the complexity can also be affected by the structural heterogeneity, since heterogeneity makes on-device optimization more difficult.
For example, although mobilenetV2 and mnasnet have similar FLOPs (300 million vs. 315 million) and parameters (3.4 million vs. 3.9 million), their performances vary a lot across DL libs.
As shown in Figure~\ref{fig:model}, squeezenet runs faster than mobilenetV2 with \snpe, \pytorchmobile, while mobilenetV2 runs faster with other DL libs.

 \implication{
The best-performing DL lib is highly fragmented across models and hardware.
Such fragmentation may even overwhelm the model designs and hardware capacity improvement.
To pursue the optimal performance in a mobile DL app, the developers need to incorporate different DL libs and dynamically load one based on the current model and hardware platform.
Such a methodology is rarely seen in practice as it incurs significant overhead to both software complexity and developing efforts.
A more lightweight system is desired to bring together the best performance of different DL libs.
}
	\subsection{Impacts of Quantization}\label{sec:quantization}

	Quantization has become a common practice to deploy DL models on mobile devices.
	There are different levels of quantization, e.g., FP16, INT16, INT8, etc~\cite{vanhoucke2011improving,gupta2015deep}.
	Among them, INT8-based quantization is known to achieve the best trade off among model accuracy and on-device speedup.
	Therefore, we mainly study INT8-based performance on CPU/GPU/DSP.
	
	\textbf{Benefit brought by INT8 quantization is under expectation.}
	Figure~\ref{fig:processor_compare} summarizes the best inference performance across DL libs on different model representations and hardware.
	It shows that quantization indeed brings inference speedup in most scenarios.
	However, the speedup (0.8$\times$--3.0$\times$) is much less than the theoretical expectation (4$\times$ due to the NEON support in Android~\cite{jo2014opencl}).
	In certain cases, the INT8-based inference is even slower than FP32, e.g., with squeezenet and vgg16 on M11 CPU.
	Furthermore, whether quantization can accelerate model inference also relies on the underlying hardware, i.e., the SoCs and the processor.
	
	We dive into the source code of those DL libs and identify the following reasons.
	(1) Modern mobile SoCs also have good support for FP processing.
	(2) FP32-based tensor operations are better tunned than INT8, according to our observations to the commit history of those DL libs.
	(3) Overhead of converting between INT8 and FP32 can incur nontrivial overhead. For example, re-quantization is essential in the final softmax layer of most classification models.

	\implication{
	Not every model can be accelerated through INT8 quantization, and the situation may vary across different hardware devices and processors.
	There exists great potential at software level to accelerate the inference of quantized models.
	}
	\subsection{Impacts of Hardware}\label{sec:accelerator}

	We then investigate whether and to what extent can more powerful CPUs or heterogeneous processors (GPU/DSP) on smartphones can accelerate DL inference.
	The results are shown in Figure~\ref{fig:processor-compare}.

\begin{figure}[t]
		\centering
		\begin{minipage}[b]{0.46\textwidth}
			\includegraphics[width=1\textwidth]{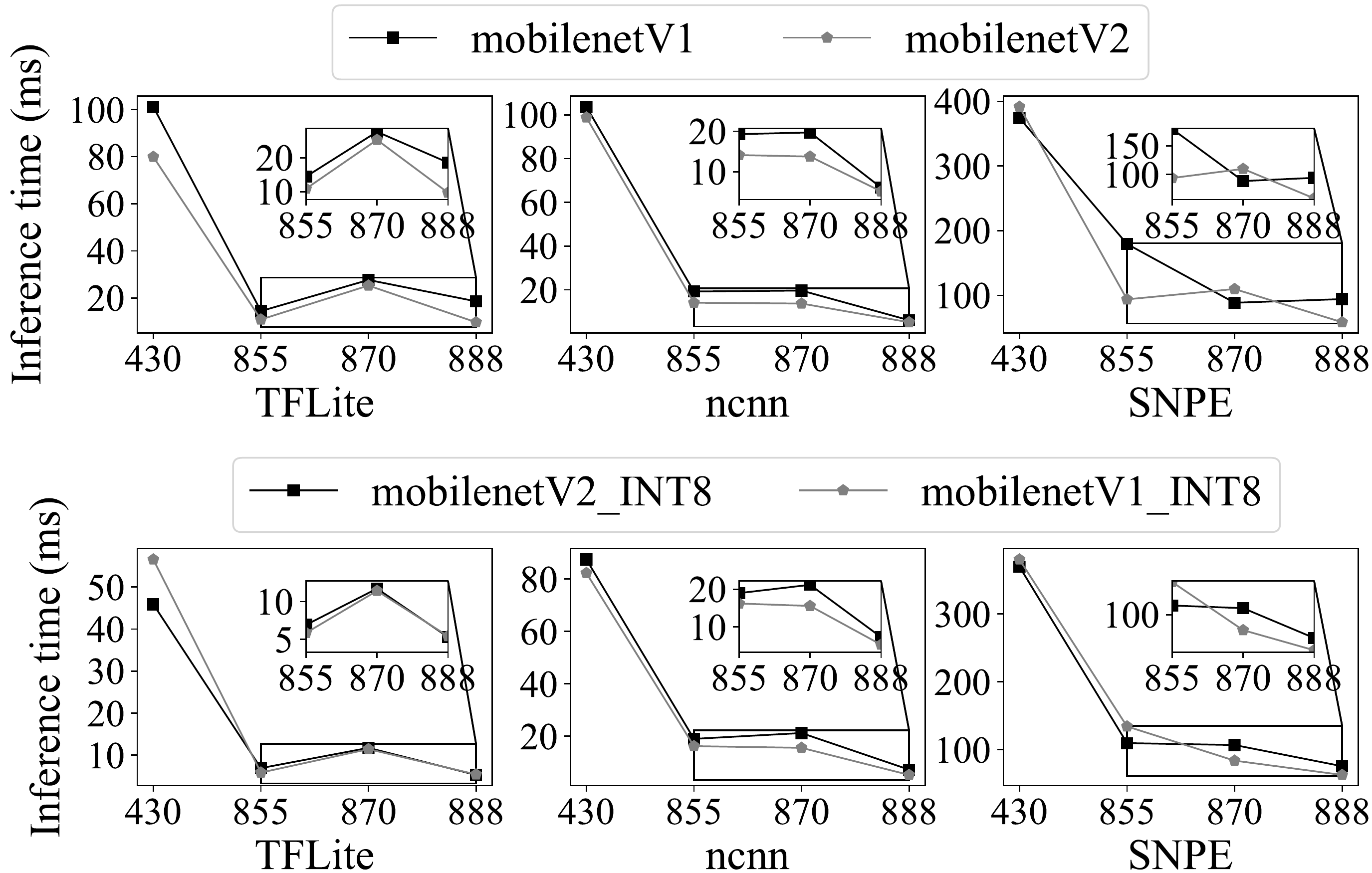}
		\end{minipage}
		\caption{The performance across different SoCs.}
		
		\label{fig:SoC_change}
		\vspace{-15pt}
	\end{figure}

	\textbf{Newer generations of mobile SoCs can mostly accelerate the inference, yet not in every case.}
	As the most representative SoC series of mobile devices, new generation of Qualcomm Snapdragon comes out every one or two years.
	As shown in Figure~\ref{fig:SoC_change}, from the Snapdragon 430 to 888, the overall performance of the three libraries (\tflite, \mnn, \snpe) shows a similar trend of improving.
	However, there are cases when newer SoC runs slower than the old ones, e.g., Snapdragon 870 vs. 855 on \tflite, even though 870 is equipped with stronger CPU and faster memory access speed~\cite{snapdragon-870-855}.
	This is mainly because Snapdragon 855 is a more popular SoC for which the DL libs are highly optimized.

	\textbf{GPUs can not always accelerate DL inference.}
	For most cases of FP32-based models, GPU can indeed bring inference speedup by 1.4$\times$--1.9$\times$ compared to CPU.
	However, in certain cases like mobilenetV1 and vgg16 on MI11, GPU even runs slower than CPU (up to 2.3$\times$).
	On INT8-based models, GPU can hardly bring any benefit.
	
	There are following primary reasons.
	Firstly, mobile GPUs are mainly designed for rendering instead of general-purpose computing.
	Their computing power is highly constrained due to the battery life consideration~\cite{chetoui2021workload}.
	Secondly, the DL libs are not as well optimized for GPUs as CPUs.
	During experiments, we observe that the arithmetic processing units inside GPU cores are often underutilized.
	Thirdly, mobile GPUs often do not have native support for INT8 data format, therefore the actual inference falls back to FP32.
	Fourth, there lack GPU support for some operators (e.g., SQUEEZE on TFLite), and those operators will fall back to run on CPUs, which incurs nontrivial overhead for data copy among CPU and GPU\footnote{Though mobile CPU and GPU share the same memory unit, their memory spaces are separated by OS and cannot be accessed mutually.}.
	\implication{
	Our findings motivate DL lib developers to focus on GPU-side optimization ~\cite{wang2021asymo}, including supporting more types of operators and single-op performance.
	It also motivates DL researchers to design the models suitable for GPU computing, that is, the operators in the models with a large number of parallel features as much as possible, and reduce high memory access operators that are not good for parallel operations.
	}

	\textbf{DSP can significantly accelerate INT8 model in most cases.}
	Figure~\ref{fig:processor-compare} also shows that running on mobile DSP can reduce the inference time of INT8 model by 2.0$\times$--12.9$\times$.
	This is mainly because Qualcomm DSP has been equipped with AI capabilities such as HTA and HTP~\cite{Qualco}, which are integrated with Hexagon vector extension (HVX) acceleration.
	Meanwhile, the Winograd algorithm is used to accelerate the convolution calculation on DSP.
	In fact, the energy saving of DSP is even more significant than inference speed (not shown in the Figure) according to our measurements.
	
	However, there are a few cases that DSP performs worse than CPU (mobilenetV1/V2 on MI11).
	This is mainly because MI11 uses Snapdragon 888 SoC, which is a relatively new chip that the DL libs are not currently well tuned for.
	
	\implication{In most cases, more powerful CPUs and accelerators (GPU and especially DSP) can speed up the model inference.
	However, there are cases that DL libs perform even worse on those hardware.
	In other words, the current DL libs can not fully exploit the capacity of each hardware.
	Our findings motivate DL lib developers to focus on optimization on heterogeneous processors~\cite{wang2021asymo}, including supporting more types of operators and single-op performance.
	It also motivates DL researchers to design the models suitable for GPU computing and reduce high memory access operators that are not good for parallel operations.
	}

	\subsection{Operator-level Integration of DL Libs}\label{sec:optimal}
	\begin{table}[t]
		\small
		\centering
		\begin{tabular}{|c|r|r|r|r|r|}
		\hline
		\textbf{ Models} & \textbf{  Mace}   & \textbf{  tflite  } & \textbf{ SNPE }  & \textbf{ ncnn} & \textbf{Oracle time}\\ \hline
		mobilenetV1                                           & 19     & 18.3  & 50.3 & 14.4  & 13.5  ($\downarrow$6.1\%)\\ \hline
		mobilenetV2                                           & 37.9  & 31.5  & 113.4 & 14.4  & 10.6 ($\downarrow$26.3\%)\\ \hline
		inceptionV3                                           & 230    & 154.9 & 176.7 & 123    & 86.3 ($\downarrow$29.9\%)                \\ \hline
		inceptionV4                                           & 293    & 195.1 & 312.6 & 374.9 & 180.3 ($\downarrow$7.6\%)                \\  \hline
		vgg16                                                 & 180.3 & 73.1  & 341.7 & 409.0 & 73.1 ($\downarrow$0\%) \\  \hline              
		\end{tabular}
		\caption{The benefits of an ``oracle DL lib'' that integrates the best-performing operator of all tested DL libs. The numbers in parenthsis indicates the potential improvement over the best-performing DL lib (different for each model).}
		\label{tab:optimal}
		\vspace{-15pt}
		\end{table}
	
	Motivated by the severe fragmentation of DL libs on diverse models and hardware, we then explore the idea of ``how much benefits can be brought if we can integrate operator-level wisdom from different DL libs''.
	More specifically, we collect the per-operator inference time for each DL lib, and combine the best-performing DL lib for each operator.
	Therefore, we obtain an ``oracle lib'' that combines the fastest operator from those DL libs.
	
	Table~\ref{tab:optimal} summarizes the performance that can be achieved by such oracle lib.
	In summary, by integrating the operator implementation of different DL libs, we can achieve 0\%--29.9\% inference time reduction compared to the best-performing DL lib across different models.
	Such improvement is nontrivial as the best-performing DL lib is already highly optimized for specific models and hardware, and their performance has reached a stable point in recent two years as shown in $\S$\ref{sec:long}.
	
	We use inceptionV4 as an example to show how different DL libs perform at operator level.
	We find \mace has best support for convolution with 3x3 kernels, while \ncnn performs best for convolution with 1x7 and 1x3 kernels.
	For some activation and software layers, \tflite performs best.
	
	\implication{
	Our findings further highlight the fragmentation of DL libs at operator level, and motivate that by combining the wisdom from different libs, the DL inference performance can be further upgraded.
	However, achieving such potential is not easy, as differnt DL libs have different ways to implement operators, e.g., the tensor alignment and memory pool.
	Those diversities need to be unified before the operator implementation can be combined.
	}

	\subsection{Cold-start Inference}\label{sec:cold-start}
	\begin{figure}[t]
		\centering
		\begin{minipage}[b]{0.46\textwidth}
			\includegraphics[width=1\textwidth]{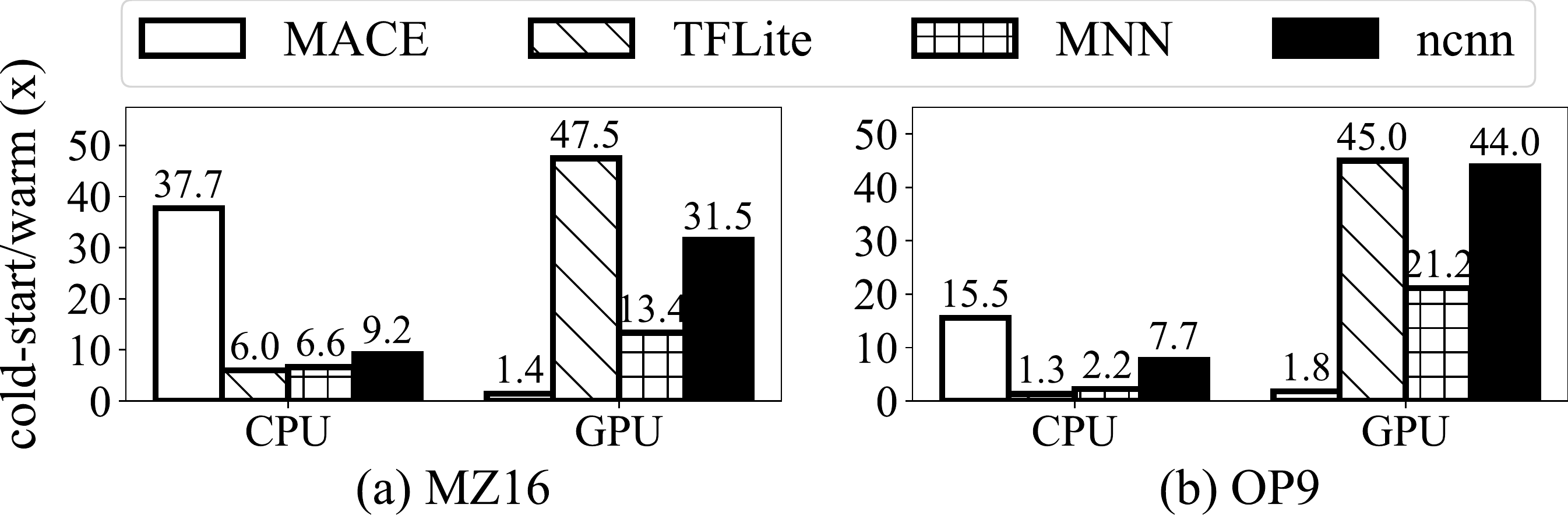}
		\end{minipage}
		
		\caption{The ratio of cold-start inference to warm inference. Numbers are averaged across all DL models.}
		
		\label{fig:warmvscold}
		\vspace{-15pt}
	\end{figure}
	
	The above results are all based on ``warm'' execution, i.e., the continuous inference after the first 5 rounds of inference.
	However, ``cold-start'' inference, i.e., the first inference beginning from model loading, is also important because for many apps the inference only happens once.
	In addition, cold-start inference is also important when apps expectedly crash and need to recover the DL functionality as fast as possible.
	
	\textbf{Cold-start inference is significantly slower than warm inference.}
	Figure~\ref{fig:warmvscold} shows how much times ({$\times$}) slower cold-start inference is on CPU and GPU averaged across all models on two mobile devices.
	Overall, cold-start inference is much slower than warm inference, i.e., 1.3$\times$--37.7$\times$ on CPU and 1.4$\times$--45.0$\times$ on GPU.
	
	\textbf{Memory preparation contributes to the largest overhead in cold-start inference.}
	To investigate the reasons of slow cold start, we dive into the source code of \ncnn and identify the workflow of the cold-start inference.
	It consists of three major steps: loading model from disk, memory preparation, and running inference.
	The memory preparation main refers to expanding the loaded weights to proper memory locations and reserving memory for intermediate feature maps to speed up the later inference.
	For example, both img2col~\cite{li2020efficient} and Winograd~\cite{wu2021exploring} implementation of convolution operation require to transform the original convolution kernel matrix to a different shape.
	
\begin{figure}[t]
		\centering
		\begin{minipage}[b]{0.48\textwidth}
			\includegraphics[width=1\textwidth]{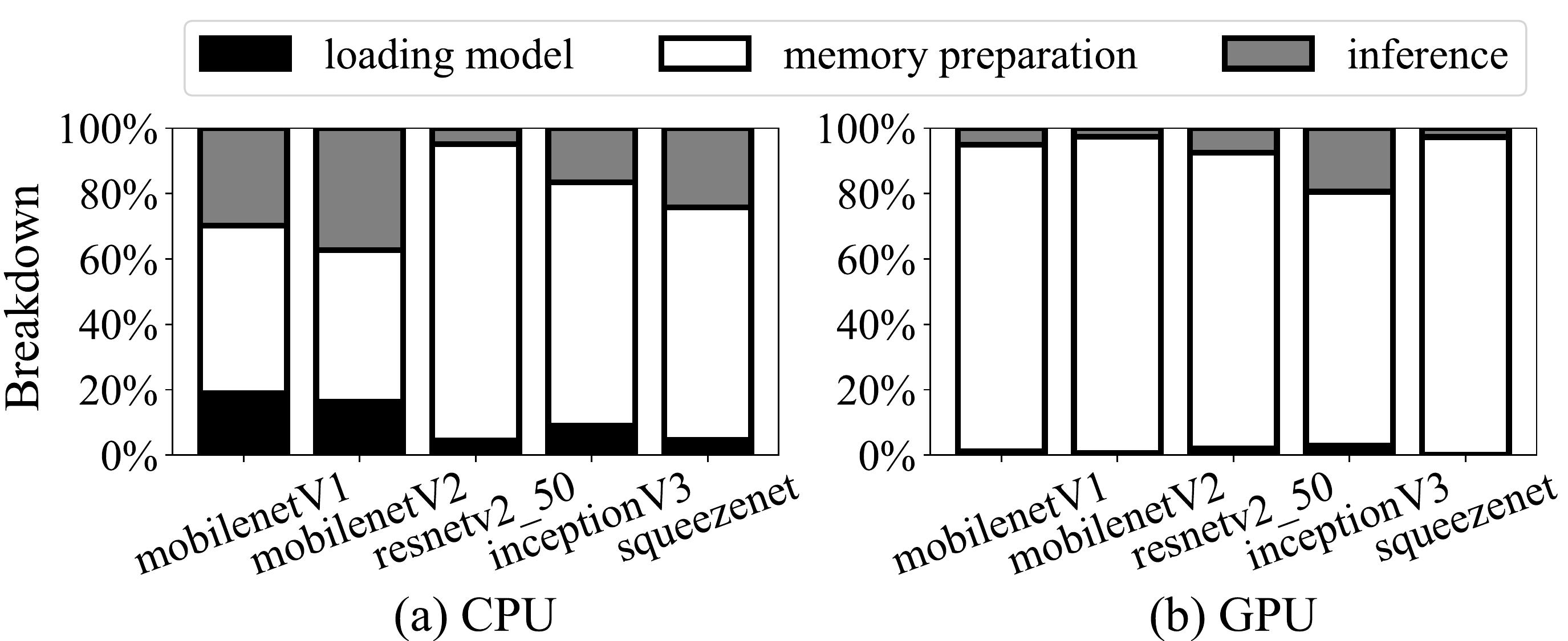}
		\end{minipage}
		
		\caption{The breakdown of cold-start inference time.}
		
		\label{fig:cold_breakdown}
		\vspace{-15pt}
	\end{figure}

	Figure~\ref{fig:cold_breakdown} quantitatively shows the breakdown of cold-start inference of \ncnn on 5 models and 2 devices.
	As observed, memory preparation is the one that accounts for the largest proportion of cold-start inference of all models, i.e., 67\% on CPU and 91\% on GPU on average.
	In fact, we observe that memory preparation is implemented in a single thread in \ncnn and other DL libs, therefore cannot benefit from the multi-core system of mobile SoCs.
	Additionally, memory preparation for GPU inference even takes more time than on CPU because of the complicated model, i.e., the code needs to be compiled to shader before executing on GPU~\cite{tornai2021compute}.

	\label{sec_design_coldstart}

	\implication{
	Optimization of cold-start inference is a rarely explored topic, but can be important in many apps that only need to execute model once each time.
	Potential solutions include speeding up memory preparation using multiple threads and generating pipeline to run model loading (I/O-intensive), memory preparation (memory-intensive), and inference (compute-intensive) simultaneously.
	}
	\subsection{Longitudinal Analysis}\label{sec:long}

	\begin{figure}[t]
		\centering
		\begin{minipage}[b]{0.44\textwidth}
			\includegraphics[width=1\textwidth]{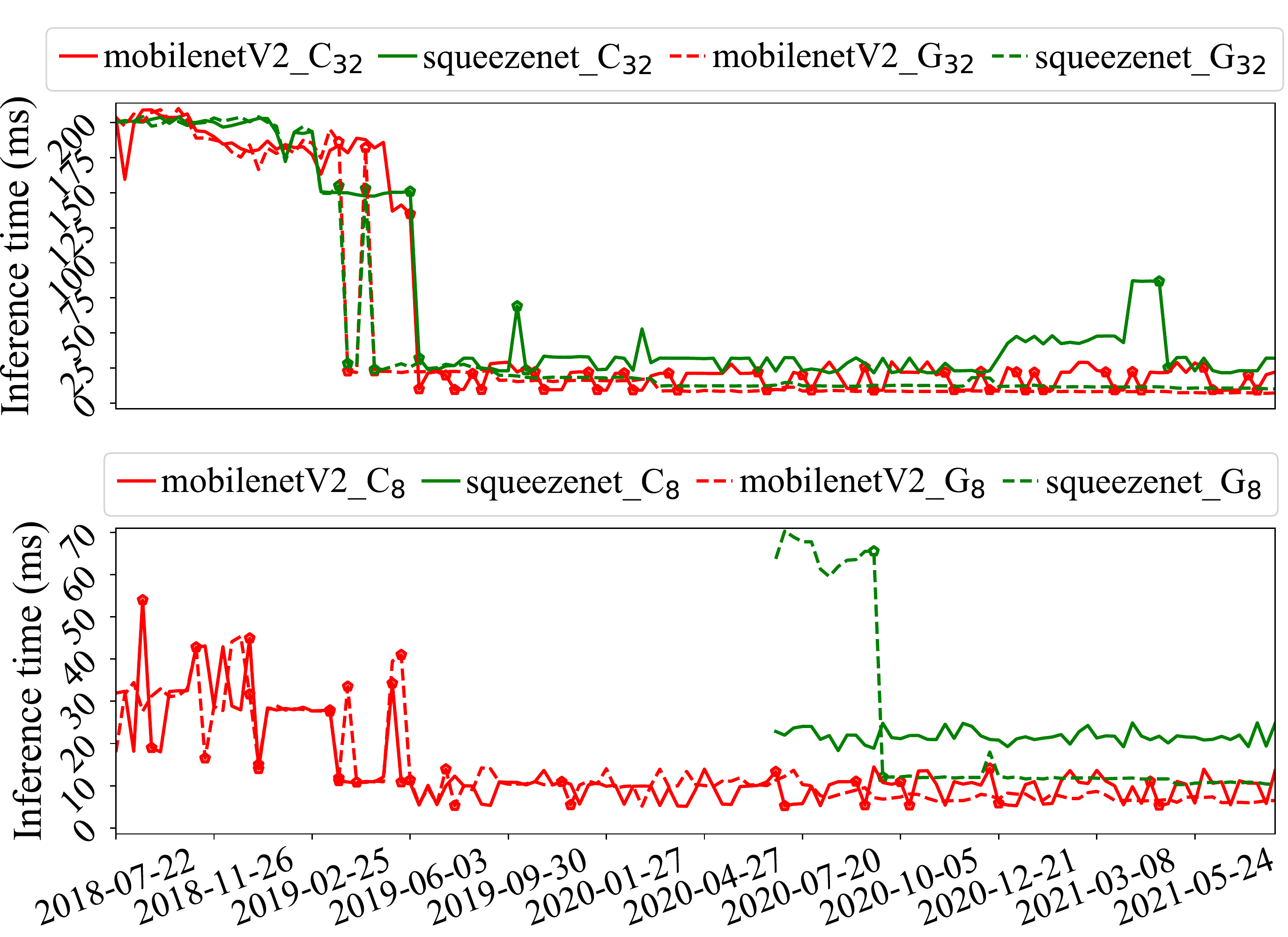}
			\subcaption{TFLite}
		\end{minipage}
	
		\begin{minipage}[b]{0.44\textwidth}
			\includegraphics[width=1\textwidth]{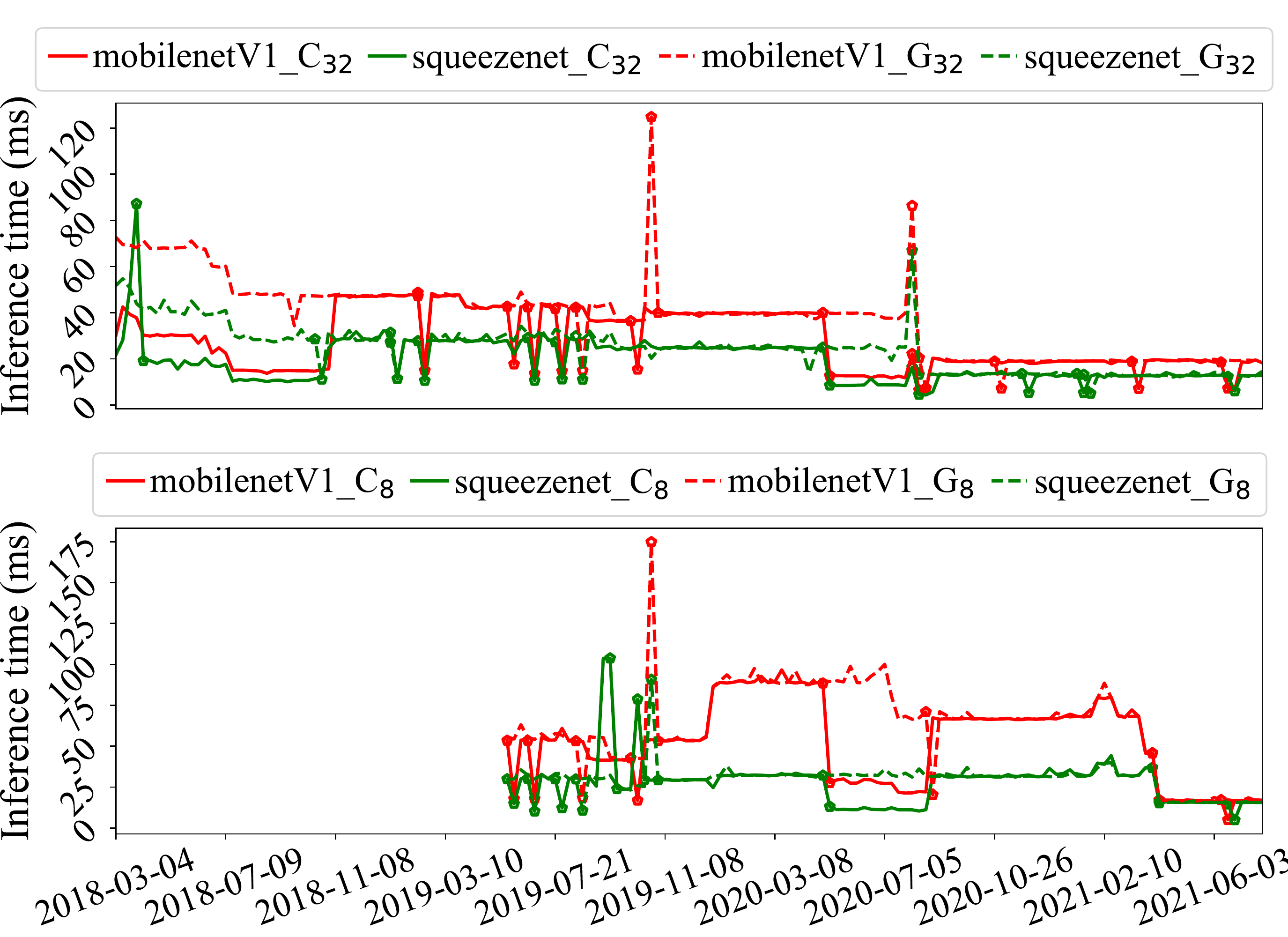}
			\subcaption{ncnn}
		\end{minipage}
		
		\caption{The inference performance evolvement across time of TFLite \& ncnn tested on HM device.}
		
		\label{fig:lifetime}
		\vspace{-15pt}
	\end{figure}
	
	We then longitudinally analyze how the performance of DL libs evolves across time.
	We select 2 DL libs that have the longest open source history and test their performance on the commits at the beginning of every week from Mar./Jul. 2018 to Jul. 2021 (80,637 commits in total) respectively.
	For simplicity, we only show test models (mobilenetV1/V2 and squeezenet) on CPU and GPU.
	
	\textbf{Overall, the performance of DL libs are continuously improving in early years, but becomes relatively stable since 2020.}
	As shown in Figure~\ref{fig:lifetime}, the performance of \tflite and \ncnn are improving: taking mobilenetV2 (FP32 format) as an example, its inference time on CPU/GPU has reduced from 203.6ms/203.8ms to 21.9ms/7.2ms with \tflite, and 30.3ms/72.7ms to 19.5ms/19.7ms with \ncnn, respectively.
	Similar observation is also made on squeezenet and the INT8 models.
	The performance improvement is mostly a cliff-like change in a few commits, rather than a regular and slow change.
	However, since 2020, the performance of DL libs is relatively stable and there are very few nontrivial improvements.
	It indicates that the DL lib community is shifting their focus from performance optimization to other aspects, e.g., supporting more types of operators.
	
	We also observe that a commit may only improve the performance of certain models.
	For example, the \emph{20275fe} commit on \tflite in Jun. 6, 2019 reduces the inference time of mobilenetV2 by 13.6$\times$, but hardly affects the inference time of mobilenetV2.
	The reason of such ``partial improvement'' is the same as the fragmentation of DL libs as mentioned in $\S$\ref{sec:fragmentation}.

	\textbf{In certain commits, we observe significant performance degradation.}
	For instance, the \emph{20cd7182} commit on \tflite in Jul. 8, 2020 increases the inference time of from 9.3ms to 21.9ms.
	Similar phenomenon also exists in \ncnn: the \emph{971fe2f} commit on \ncnn in Nov. 3, 2018 increases the inference time of squeezenet from 11.6ms to 24.8ms.
	We regard such commits that cause significant performance degradation on certain models and devices as \textit{performance bug}.
	We dig into those commits' contents, and find that except ``real bugs'', a common reason is that a new operator implementation is pushed to improve the performance of DL models, but do not work well under certain settings, which is unexpected to the commit submitter.
	For instance, a convolution kernel may perform well with 3x3 kernel size and stride size 2, but not with 3x3 kernel size and stride size 1.
	Among the 34 detected performance bugs, it takes around 1--16 weeks to fix so that the performance can be recovered.
	
	\implication{
	The current open-source ecosystems of DL libs sometimes introduce performance bugs, possibly due to a comprehensive benchmarking tool available for developers to test their commits.
	Indeed, due to the performance heterogeneity of DL libs on different models and hardware, it is almost impossible to fully eliminate performance bugs.
	We propose two possible solutions.
	One is to set up an environment with diverse device models periodically (e.g., per day) running a comprehensive benchmark like \sys to timely detect performance bugs.
	Another one is to build a static analysis tool that can identify commits with potential bugs based on history.
	}
	\section{Related Work}\label{sec:related}

\noindent \textbf{Mobile DL} In recent years, there is a notable trend to move DL inference into local devices instead of offloading to remote servers~\cite{deeptype,xu2018deepcache,leontiadis2021s,laskaridis2020spinn,diva,elf,deepwear}.
A fundamental challenge of this trend is the constrained resources of smartphones.
Therefore, performance optimization has been a primary research direction for both academia and industry~\cite{tensorflow,pytorch,laskaridis2021adaptive,yeo2020nemo,zhang2020mobipose,op,aliedge}.
There have been some optimization research efforts addressed to recude the overhead of DL on smartphones,
e.g., offloading, model quantization and sparsity~\cite{wu2020emo,kang2017neurosurgeon,jacob2018quantization,liu2018demand}. These solutions usually either count on preprocessing or perform under lab simulations on the data collected
preciously from smartphones.
Thus, our work brings DL to smartphones in the real world and provides a unified
approach to easily compare performance among different libs. This work is motivated by many years of efforts at this lane.

\noindent \textbf{AI benchmarks}
As summarized in Table~\ref{tab:DL_libs}, there exist a few AI benchmarks for diversified scenarios, e.g., datacenter servers or edges, inference or training,
 etc~\cite{AIIA,DAWNBench,DeepBench,adolf2016fathom,almeida2021smart,ignatov2019ai,mattson2019mlperf}.
This work explicitly targets at inference on mobile devices.
Besides, the ecosystem of on-device DL libs is more fragmented than servers due to the high fragmentation of mobile hardware.
Furthermore,
a number of studies focus on DL libs analysis.
Consequently, the results are limited in small-scale project from the specific perspective.
Luo et al.~\cite{luo2020comparison} proposed the benchmark suite for evaluating the abilities of mobile devices across different libs.
MLPerf~\cite{mattson2019mlperf} proposed high-level rules for more flexible benchmark of the libs.
Tang et al.~\cite{tang2021bridge} studied the behavior characteristics of neural networks to bridge networks design and real-world performance.
There is still limited understanding about the performance of DL libs across heterogeneous smartphones.
Compared to similar benchmarks focusing on DL libs, \sys has richer support for various DL libs and models.

\noindent \textbf{Empirical study of mobile DL}
One line of studies mainly focus on DL apps/systems/models.
Xu et al.~\cite{xu2019first} demystified how smartphone apps exploit DL models by deeply analyzing Android apps.
Wang et al.~\cite{wang2019understanding} made efforts towards the evolution of mobile app ecosystem.
Andrey et al.~\cite{ignatov2018ai} targeted at devices and focused on running models with hardware acceleration of smartphones.
Although the studies have analyzed on device DL, they lack a comprehensive understanding and benchmarking on diverse libs.

\noindent \textbf{Deep learning compilers}~\cite{dalibard2017boat,ahn2020chameleon,zheng2020ansor}
represent a different way to deploy DL models compared to static libs.
The key idea of DL compilers is to pre-define primitives of operators and rules to find an optimal implementation.
Our benchmark does not include such compilers because of following reasons.
First, the state-of-the-art DL compilers like TVM~\cite{chen2018tvm} are mainly designed for servers instead of mobile devices.
According to our measurements, on most of models, the resultant performance is not even close to our tested DL libs after many hours of search.
Second, because of the high fragmentation of mobile devices, generating an execution plan for each device is impractical.
In fact, some DL libs already search for an optimal working group size in their GPU implementation.

\section{Conclusions}\label{sec:conclusions}

In this work, we built the first comprehensive benchmark for DL libs and conducted extensive measurements to quantitatively understand their performance.
The results help reveal a complete landscape of the DL libs ecosystem.
Atop the observations, we summarize strong implications that can be useful to developers and researchers.
\section*{Acknowledgments}
This work was partly supported by National Key R\&D Program of China under grant number 2020YFB1805500, and National Natural Science Foundation of China under grant number 61922017 and 62102009.
Mengwei Xu was sponsored by CCF- Baidu Open Fund.
Xuanzhe Liu was supported by PKU-Baidu Fund Project under the grant number 2020BD007 and Alibaba Group through Alibaba Innovative Research (AIR) Program.
	\bibliographystyle{plain}
	\bibliography{ref_mwx}

\end{document}